\documentclass[10pt,twocolumn,letterpaper]{article}

\usepackage{cvpr}
\usepackage{times}
\usepackage{epsfig}
\usepackage{graphicx}
\usepackage{amsmath}
\usepackage{amssymb}
\usepackage{mathtools}
\usepackage{comment}
\usepackage{caption}
\usepackage{subcaption}

\newcommand{\aacomment}[1]{\noindent{\textcolor{blue}{\textbf{\#\#\# AA:} \textsf{#1} \#\#\#}}}

\usepackage[pagebackref=true,breaklinks=true,letterpaper=true,colorlinks,bookmarks=false]{hyperref}

\cvprfinalcopy 

\def\cvprPaperID{****} 

\ifcvprfinal\fi
\begin{document}

\title{Compact Tensor Pooling for Visual Question Answering}

\author{Yang Shi\\
University of California, Irvine\\
{\tt\small shiy4@uci.edu}
\and
Tommaso Furlanello\\
University of Southern California\\
{\tt\small tfurlanello@gmail.com }
\and
Anima Anandkumar\\
Amazon\\
{\tt\small anima@amazon.com}
}

\maketitle

\begin{abstract}
\vspace*{-10pt}
Performing high level cognitive tasks requires the integration of feature maps with drastically different structure. In Visual Question Answering (VQA) image descriptors have spatial structures, while lexical inputs inherently follow a temporal sequence. The recently proposed Multimodal Compact Bilinear pooling (MCB) forms the outer products, via count-sketch approximation, of the visual and textual representation at each spatial location. While this procedure preserves spatial information locally, outer-products are taken independently for each fiber of the activation tensor, and therefore do not include spatial context.
In this work, we introduce \emph{multi-dimensional sketch} ({MD-sketch}), a novel extension of count-sketch to tensors. Using this new formulation, we propose Multimodal Compact Tensor Pooling (MCT) to fully exploit the global spatial context during bilinear pooling operations.
Contrarily to MCB, our approach preserves spatial context by directly convolving the MD-sketch from the visual tensor features with the text vector feature using higher order FFT. Furthermore we apply MCT incrementally at each step of the question embedding and accumulate the multi-modal vectors with a second LSTM layer before the final answer is chosen.

\end{abstract}
\vspace*{-20pt}
\section{Introduction}
\vspace*{-6pt}
Visual question answering tasks require a joint representation of both visual and textual input~\cite{paper:VQA}. Image and text features are respectively extracted with convolutional neural networks \cite{DBLP:journals/corr/DonahueJVHZTD13} and recurrent neural networks \cite{DBLP:journals/corr/SutskeverVL14, hochreiter1997long}. 
Information of each modality are then combined together, using concatenation, element-wise product or sum operations. 
Bilinear pooling\cite{CBP} exploits the outer product operation to capture richer information from the two sets of features. 
However, its computational/memory complexity (O($n^2$), if each feature is $O(n)$) is prohibitive in practice.
Compact bilinear pooling \cite{tensorsketch} has the same discriminative 
power as the bilinear pooling 
while requiring far less computation/memory storage (O($d$), $d \ll n^2$). 

In this paper, we first propose the MD-sketch that generalizes count-sketch to tensors of arbitrary orders. We then leverage this formulation of the MD-sketch in a new compact tensor pooling method.
This new pooling, which we coin \emph{MCT}, directly estimates the outer product between tensor descriptors of different orders, therefore preserving context. We present novel VQA architectures with MCT and also aggregate MCT within LSTM. Specifically, we apply the MCT model at each time-step of the language LSTM to combine the image features with each word representation.
\vspace*{-10pt}
\section{Related work}
\vspace*{-6pt}
\textbf{Tensor sketch} 
Count Sketch~\cite{countsketch} estimates the data frequency in a stream. Considering that as a specific random projection in high dimensional space, Pham and Pagh propose tensor sketch for polynomial kernel estimation in~\cite{tensorsketch}. Wang et al.~\cite{tensordecompose} computes tensor contraction via tensor sketch. The computational burden is reduced while performing the operations by sketching the tensor as a vector. However, this ignores the natural data structure.
We generalize count-sketch method to tensors of arbitrary orders while preserving spatial context and avoiding expensive computation in Section~\ref{sec:MCT}.

\textbf{Pooling methods}
Pooling methods are widely used in visual tasks to combine information for various streams into one final feature representation. Common pooling methods are average pooling and bilinear pooling. 
Bilinear pooling requires taking the outer product between two features. However the result is high dimensional. Count sketch is applied as a feature hashing operator to avoid such increase in dimension. Gao et al.~\cite{CBP} uses convolution layers from two different neural networks as the local descriptor extractors of the image and combine them using count sketch. Simon et al.~\cite{alphapooling} presents a new "$\alpha$-pooling" which allows the network to learn the pooling strategy: a continuous transition between linear and polynomial pooling. They show that higher $\alpha$ gives larger gain for fine-grained image recognition tasks. However, as $\alpha$ goes up, the computation complexity increases in polynomial order. We propose up with compact polynomial pooling to reduce the computational burden. Details in Section~\ref{sec:MCB}.
\vspace*{-10pt}
\section{Multimodal compact bilinear pooling(MCB)}
\vspace*{-6pt}
\label{sec:MCB}
Fukui et al.~\cite{MCB} uses count sketch as a pooling method in VQA tasks and obtains the best results on VQA dataset v1.0.
We denote the count sketch operator as $cs$, the tensor product (outer product in vector case) as $\otimes$ and convolution operator as $\star$. The key idea in MCB is to avoid directly computing the outer product and to reduce the output dimension by writing $cs(x \otimes y) = cs(x) \star cs(y)$. According to convolution theorem, this can be further simplified by doing element-wise product in the frequency domain before transforming the result back to the time domain using inverse FFT(IFFT). Here, we argue that the IFFT step is redundant in the VQA task. FFT itself preserves the correlation between the two data streams. We can therefore accelerate the training by skipping the IFFT step. By padding the representation vectors with $\vec{1}$(vector with all elements 1) before the count sketch, we additionally preserve the first order information along with the joint correlation. Assume we have two input features $x,\vec{1}_2 \in \mathcal{R}^{n_1}$, $y, \vec{1}_1 \in \mathcal{R}^{n_2}$, set $\tilde{x} = [x,\vec{1}_1]$, $\tilde{y} = [y,\vec{1}_2]$ such that $\tilde{x}, \tilde{y} \in \mathcal{R}^{n_1+n_2}$,  $cs(\tilde{x} \otimes \tilde{y})$ will contain information of $x \otimes y$, $x$ and $y$. By repeating the same operation multiple times it is possible to obtain estimates of higher polynomials of $x$ and $y$.
\vspace*{-10pt}
\section{Multimodal compact tensor pooling(MCT)}
\vspace*{-6pt}
\label{sec:MCT}
While MCB shows great performance in \cite{MCB}, it loses the spatial context in the image
as the sketch is done independently between the 1-dimensional text feature vector and each fiber of the image tensor.
We instead propose to preserve that context in VQA tasks, by applying MD-sketch on both the image tensor and the text features and convolving the resulting tensors.

\begin{figure}
\centering
\begin{minipage}{.25\textwidth}
  \centering
  \includegraphics[width=3cm,height=1.5cm]{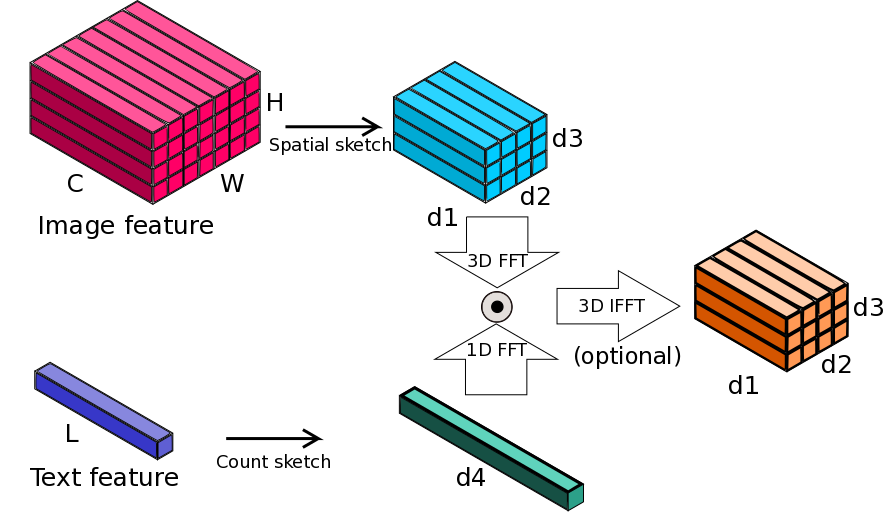}
  \vspace*{-5pt}
  \captionof{figure}{MCT in VQA}
  \label{fig:1}
\end{minipage}%
\begin{minipage}{.25\textwidth}
  \centering
  \includegraphics[width=3cm,height=1.5cm]{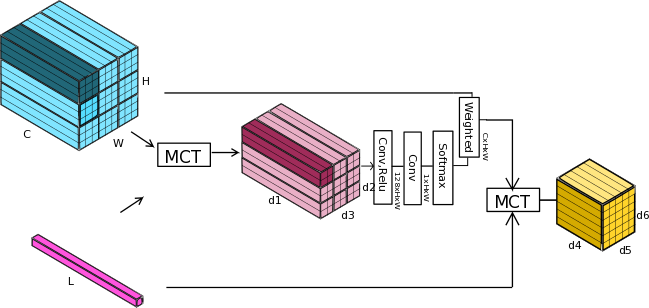}
  \vspace*{-5pt}
  \captionof{figure}{MCT-group in VQA}
  \label{fig:2}
\end{minipage}
\end{figure}

\textbf{MD-sketch of a tensor} Even though the MD-sketch is defined for tensors of arbitrary orders, we limit ourselves to third order tensors for clarity. Given a feature tensor $\mathcal{I} \in \mathcal{R}^{C \times H \times W}$, a random hash functions $h_1$:$[C]$ $\to$ $[d_1]$, $h_2$:$[H]$ $\to$ $[d_2]$, $h_3$:$[W]$ $\to$ $[d_3]$, and a random sign function $s_1$:$[C]$ $\to \{\pm 1\}$, $s_2$:$[H]$ $\to \{\pm 1\}$, $s_3$:$[W]$ $\to \{\pm 1\}$, for $i \in [C]$, $j \in [H]$, $k \in [W]$, the MD-sketch of $\mathcal{I}$ is defined as:
\vspace*{-6pt}
\begin{align*}
ss(\mathcal{I})_{t_1, t_2, t_3} & = \mathcal{X}(t_1, t_2, t_3) \\
& = \sum_{\mathclap{h_1(i)= t_1,h_2(j)= t_2,h_3(k)= t_3}}s_1(i)s_2(j)s_3(k)\mathcal{I}(i,j,k)
\end{align*}
The resulting tensor $\mathcal{X} \in \mathcal{R}^{d_1 \times d_2 \times d_3}$ is an estimation of $\mathcal{I}$ with preserved spatial information. 
We apply count sketch on text feature $v \in \mathcal{R}^{L}$. Given hash function $h_4$:$[L]$ $\to$ $[d_4]$ and sign function $s_4$:$[L]$ $\to \{\pm 1\}$:
\vspace*{-6pt}
\begin{align*}
cs(v)_{t_1}  = w(t_1)  = \sum_{h_4(i)=t_1}s_4(i)v(i)
\end{align*}
\vspace*{-4pt}
To compute the bilinear pooling between $\mathcal{I}$ and $v$, we propose MCT as follows:
\vspace*{-8pt}
\begin{align*}
MCT&(\mathcal{I} \otimes v)_{t_1, t_2, t_3} = \mathcal{Y}(t_1, t_2, t_3) \\
& = (ss(\mathcal{I}) \star cs(v))_{t_1, t_2, t_3}\\
& = 3d FFT(ss(\mathcal{I}))_{t_1, t_2, t_3} \cdot 1d FFT(cs(v))_{t_4}
\end{align*}
where $t_4 = (t_1+t_2+t_3)$ mod $d_4$.
We show the MCT module in Figure~\ref{fig:1}. 

The intuition here is that the visual representation at each grid location is not fully isolated from grids around it. When applying compact bilinear pooling along each pixel, spatial context information is lost. By combining these surrounding information using MCT, we preserve the natural tensor structure of the image. In practice, we can do local tensor spatial pooling: divide the image tensor in several sub tensors and apply MCT on each of them.

\vspace*{-10pt}
\section{VQA architecture}
\vspace*{-6pt}
\label{sec:vqa}

\begin{figure}
\centering
\includegraphics[height = 2.8cm, width = 9.5cm]{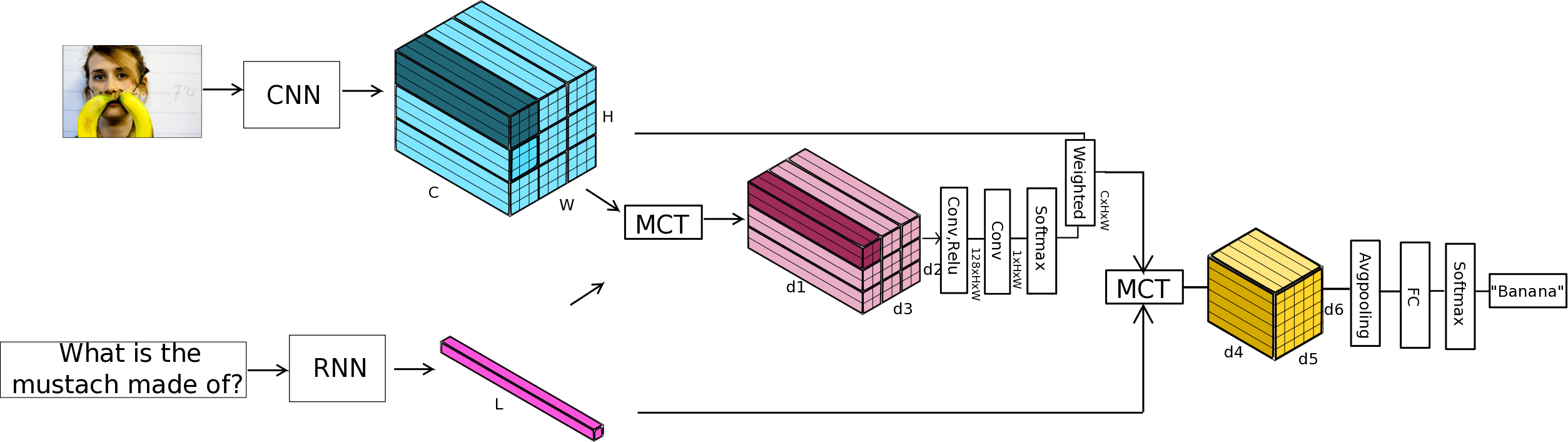}
\vspace*{-10pt}
\caption{Our architecture for VQA with MCT}
\label{fig:3}
\end{figure}

\textbf{MCT}
The first extension that we propose in Figure~\ref{fig:3} is the use of MCT instead of MCB during the feature integration steps. In the attention pathway we use MD-sketch to project the image feature to a lower dimensional space. After 2 layers of convolution we compute a soft attention map of the original image size. After filtering the image features with the attention map the resulting tensor is combined through MCT with the LSTM vector again before a final choice is taken. 

\begin{figure}
\centering
\includegraphics[height = 3cm, width = 8.5cm]{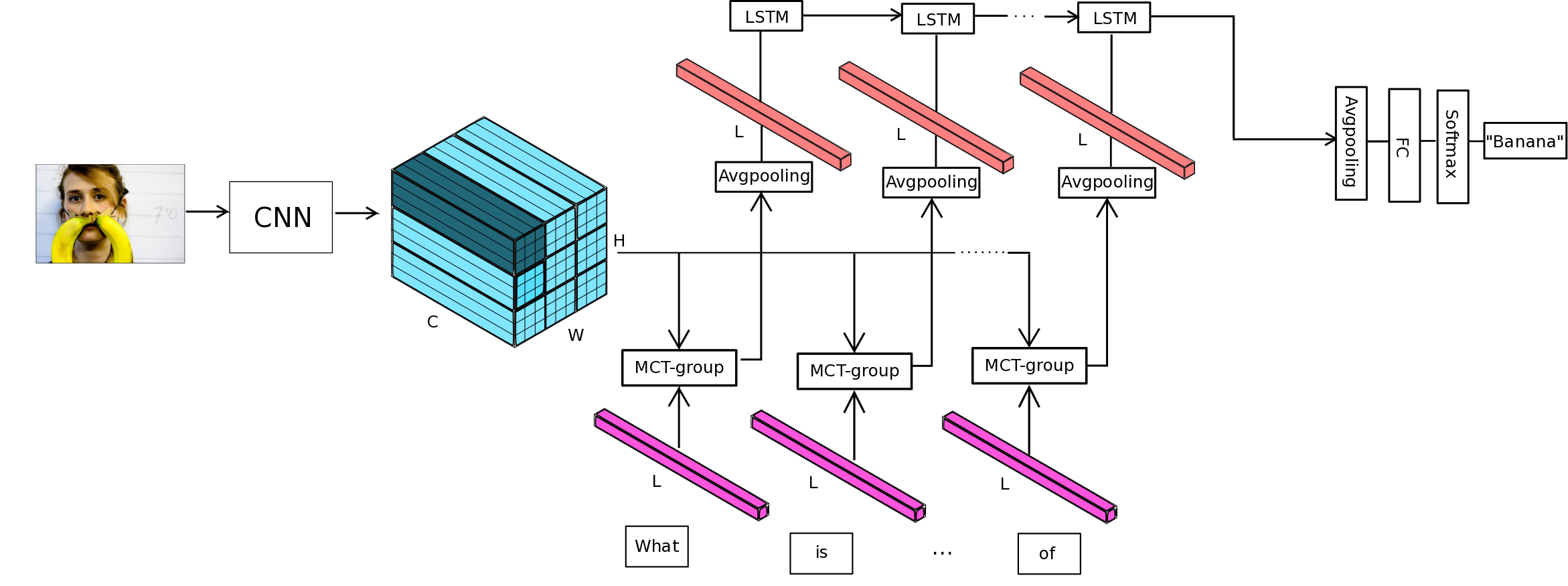}
\vspace*{-16pt}
\caption{Our architecture for VQA with LSTM-MCT, MCT-group is as per in Figure~\ref{fig:2}}
\label{fig:4}
\end{figure}

\textbf{LSTM-MCT}
We propose in Figure~\ref{fig:4} a sequential model to answer the question incrementally by giving a glimpse to the image for each word. We apply the MCT model at each time-step of the language LSTM and obtain a sequence of multimodal vectors. A second LSTM is used to aggregate and answer from the sequence of multimodal vectors.

\vspace*{-5pt}
\vspace*{-5pt}

\end{document}